\documentclass{article}

\usepackage[preprint]{neurips_2025}
\usepackage[utf8]{inputenc} %
\usepackage[T1]{fontenc}    %
\usepackage{hyperref}       %
\usepackage{url}            %
\usepackage{booktabs}       %
\usepackage{amsfonts}       %
\usepackage{nicefrac}       %
\usepackage{microtype}      %
\usepackage{xcolor}         %
\usepackage{algorithm}      %
\usepackage[noend]{algpseudocode}
\usepackage{amsmath}
\usepackage{wrapfig}
\usepackage{enumitem}
\usepackage{subfigure}
\usepackage{url}
\usepackage{enumitem}

\usepackage{hyperref}
\setlist[itemize]{noitemsep, topsep=0pt, partopsep=0pt, parsep=0pt, leftmargin=2em}

\setlist[enumerate]{noitemsep, topsep=0pt, partopsep=0pt, parsep=0pt, leftmargin=2em}

\usepackage{xcolor}
\usepackage{soul}

\definecolor{myblue}{RGB}{47,47,148}
\definecolor{mygreen}{RGB}{64,120,50}
\definecolor{myorange}{RGB}{121,68,22}
\definecolor{mypurple}{RGB}{108,40,61}

\definecolor{lightblue}{RGB}{235,245,255}
\definecolor{lightgreen}{RGB}{240,255,240}
\definecolor{lightorange}{RGB}{255,240,230}
\definecolor{lightpurple}{RGB}{250,240,255}

\newcommand{\colorhighlight}[3]{%
  \sethlcolor{#2}%
  \textcolor{#1}{\hl{#3}}%
}

\makeatletter
\renewcommand{\thefootnote}{\fnsymbol{footnote}}
\makeatother

\usepackage[font=small,labelfont=bf]{caption} 
\usepackage[subrefformat=parens]{subcaption}
\usepackage{tcolorbox}
\tcbuselibrary{listings, breakable}

\newtcolorbox{promptbox}{
    fonttitle = \bfseries, fontupper = \sffamily, fontlower = \sffamily,
    title={Prompt}
}
\newtcolorbox{rbox}[1]{
    fonttitle = \bfseries, fontupper = \sffamily, fontlower = \sffamily,
    colbacktitle = red!50!white, title={#1}
}
\newtcolorbox{bbox}[1]{
    fonttitle = \bfseries, fontupper = \sffamily, fontlower = \sffamily,
    colbacktitle = blue!50!white, title={#1}
}

\tcbuselibrary{listings, breakable} 

\newcounter{gboxcounter}
\newtcolorbox[auto counter, number within=section, number freestyle={\noexpand\arabic{gboxcounter}}]{gbox}[2][]{
    fonttitle = \bfseries, fontupper = \sffamily, fontlower = \sffamily,
    colbacktitle = teal!50!white, title={#1},
    listing only,
    breakable, 
    listing options={
        basicstyle=\tiny\ttfamily,
        breaklines=true,
        showspaces=true, 
        showtabs=true, 
        showstringspaces=true,
    },
    title={Prompt \thegboxcounter: #2},
    label={#1}, 
    before upper={\stepcounter{gboxcounter}}
}

\title{Learning Instruction-Following Policies through Open-Ended Instruction Relabeling with Large Language Models}

\author{%
\textbf{Zhicheng Zhang$^{1}$\footnotemark[1]~, Ziyan Wang$^{2}$\footnotemark[1]~~\footnotemark[2]~, Yali Du$^{2}$~, Fei Fang$^{1}$~} \\[0.5em]
$^1$Software and Societal Systems Department, Carnegie Mellon University, Pittsburgh, PA, USA\\ $^2$Department of Informatics, King's College London, London, UK\\
}

\begin{document}

\renewcommand{\thefootnote}{\fnsymbol{footnote}}
\footnotetext[1]{Equal contribution.}
\footnotetext[2]{Work done during Ziyan’s visit at CMU.}
\renewcommand*{\thefootnote}{\arabic{footnote}}
\renewcommand{\thefootnote}{\fnsymbol{footnote}}

\maketitle

\begin{abstract}
Developing effective instruction-following policies in reinforcement learning remains challenging due to the reliance on extensive human-labeled instruction datasets and the difficulty of learning from sparse rewards. In this paper, we propose a novel approach that leverages the capabilities of large language models (LLMs) to automatically generate open-ended instructions retrospectively from previously collected agent trajectories. Our core idea is to employ LLMs to relabel unsuccessful trajectories by identifying meaningful subtasks the agent has implicitly accomplished, thereby enriching the agent's training data and substantially alleviating reliance on human annotations. Through this open-ended instruction relabeling, we efficiently learn a unified instruction-following policy capable of handling diverse tasks within a single policy. We empirically evaluate our proposed method in the challenging Craftax environment, demonstrating clear improvements in sample efficiency, instruction coverage, and overall policy performance compared to state-of-the-art baselines. Our results highlight the effectiveness of utilizing LLM-guided open-ended instruction relabeling to enhance instruction-following reinforcement learning.
\end{abstract}

\section{Introduction}

Instruction-following reinforcement learning (RL), where agents learn to efficiently interpret and execute tasks specified through natural-language instructions, holds immense promise for building generalizable and flexible AI systems. Despite considerable advances in goal-conditioned RL methods~\citep{schaul2015universal,andrychowicz2017hindsight}, existing instruction-following RL approaches continue to face significant challenges. Typically, such methods heavily rely on large-scale human-annotated instruction datasets~\citep{hill2020human,narasimhan2018grounding} or predefined instruction templates, limiting scalability, generalization capabilities, and thereby constraining their real-world applicability. Furthermore, environments characterized by sparse feedback exacerbate this challenge—agents are likely to collect numerous unsuccessful trajectories that offer minimal utility, leading to inefficient exploration and slow policy improvement.

In this paper, we propose a novel framework to address these challenges by leveraging the strong reasoning capabilities of pretrained large language models (LLMs). Our key insight is to apply LLMs retrospectively to generate meaningful, open-ended instructions from collected agent trajectories. Specifically, the LLM identifies semantically relevant subtasks that the agent implicitly accomplished within failed trajectories and provides corresponding instructions, thus enriching these experiences with informative reward signals. Through this automatic instruction relabeling, we efficiently transform previously sparse and unsuccessful trajectories into valuable learning samples, thereby improving the data efficiency and diversity of instruction-conditioned policy learning without requiring any manual annotation effort.

We empirically validate our method on Craftax~\citep{matthews2024craftax}, a challenging benchmark that offers diverse semantic instructions and inherently sparse rewards. Experimental results demonstrate that our proposed framework significantly surpasses strong baselines, achieving substantial improvements in terms of sample efficiency, coverage and diversity of instructions, and the overall quality of learned instruction-following policies. Our primary contributions can be summarized as follows:
\begin{enumerate}
\item We propose a novel open-ended instruction relabeling framework that leverages large language models to automatically assign semantically meaningful instructions to collected trajectories, completely eliminating dependence on manual human annotations.
\item Our relabeling method effectively transforms unsuccessful trajectories into informative training examples, enabling more efficient learning of instruction-following policies in open-ended environments.
\item Extensive empirical experiments conducted on the Craftax benchmark validate that our approach substantially outperforms state-of-the-art instruction-conditioned reinforcement learning baselines, demonstrating strong improvements in sample efficiency, instruction diversity, and instruction-following performance.
\end{enumerate}

\begin{figure}[t]
    \centering
    \includegraphics[width=\linewidth]{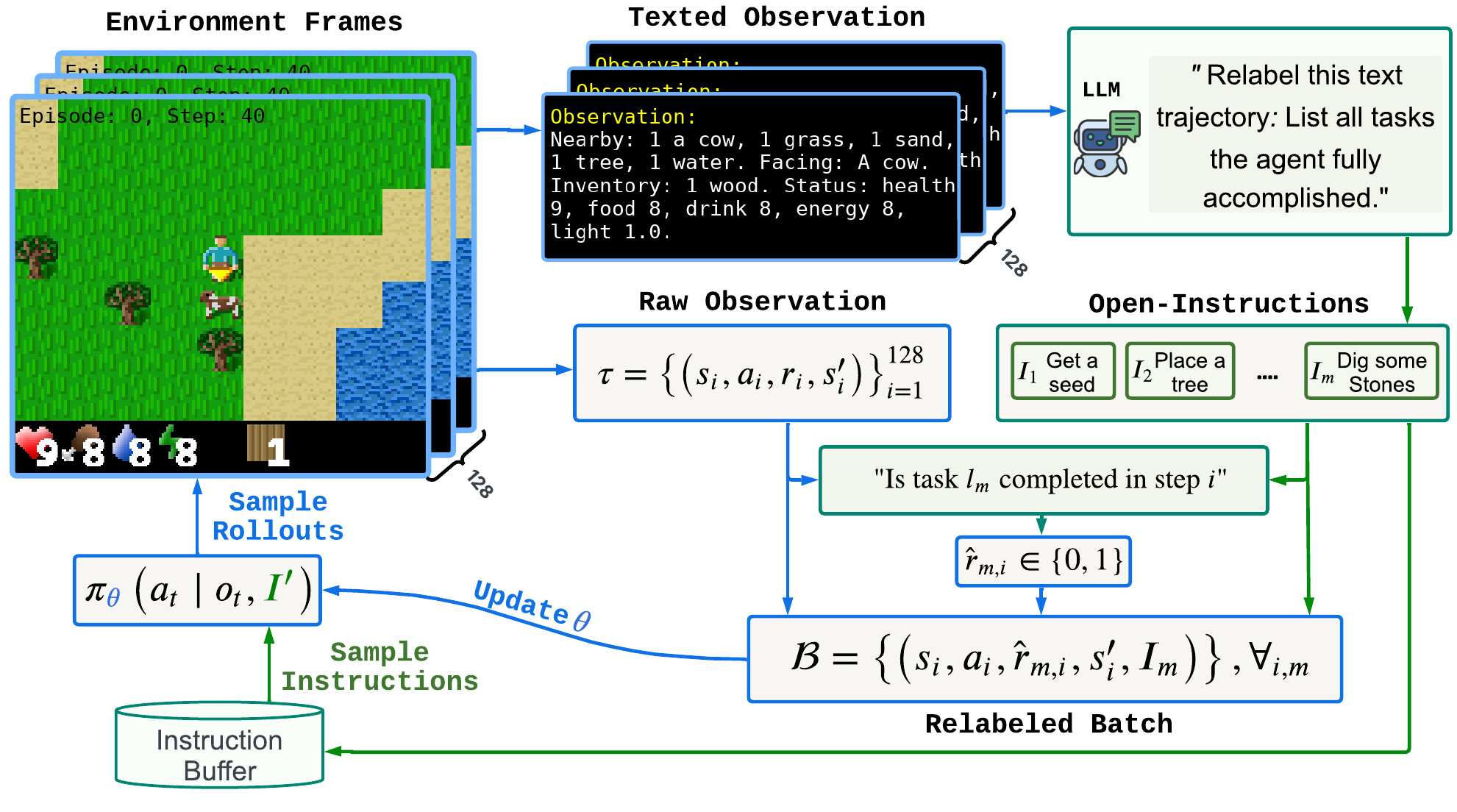}
    \caption{Overview of \textbf{OIR} framework. \colorhighlight{myblue}{lightblue}{ $\rightarrow$ \textbf{Blue flow}  $\rightarrow$} illustrates standard reinforcement learning: an instruction-conditioned policy samples rollouts from the environment using instructions sampled from the instruction buffer, after which the policy parameters are updated with the collected samples. \colorhighlight{mygreen}{lightgreen}{ $\rightarrow$\textbf{ Green flow $\rightarrow$}} highlights our novel relabeling mechanism: converting collected trajectories into extual observations and then prompting a pretrained LLM to retrospectively generate diverse, open-ended instructions identifying successfully accomplished subtasks. These generated instructions then serve to relabel trajectories by producing binary, semantic rewards, thereby enriching the instruction buffer with new and informative learning signals.}
    \label{fig:enter-label}
    \vspace{-0.7cm}
\end{figure}

\section{Related Work}

\noindent \textbf{Goal-Conditioned Reinforcement Learning (GCRL).} Goal-conditioned reinforcement learning generalizes standard RL objectives by conditioning policies and reward functions explicitly on goal representations~\citep{schaul2015universal,liu2022goal}. Many GCRL methods leverage hindsight relabeling strategies to address sparse rewards, notably Hindsight Experience Replay (HER)~\citep{andrychowicz2017hindsight}. HER-inspired methods propose various heuristics to select suitable goals, such as goal discovery based on reward relevance~\citep{pitis2020maximum,fang2019curriculum}, goal diversity~\citep{ren2019exploration}, and adaptive difficulty selection based on learning progress~\citep{nair2018visual,warde2018unsupervised}. However, existing methods primarily rely on predefined numeric or state-based goals and cannot easily scale to instructions presented in open-ended natural language.

\noindent \textbf{Instruction-Conditioned Reinforcement Learning.} Instruction-conditioned RL extends goal-conditioned RL by formulating tasks explicitly via natural language instructions that are encoded directly into policy inputs and rewards~\citep{luketina2019survey,narasimhan2018grounding,hill2020human}. Recent works propose using hindsight instruction relabeling (HIR)~\citep{zhang2023wisdom} methods that retrospectively assign appropriate instruction labels to collected trajectories, substantially enhancing data efficiency and policy robustness. Nonetheless, conventional HIR approaches typically require human-defined templates or substantial manual labeling efforts, limiting their scalability and generalization potential. Our approach, in contrast, eliminates the need for manually defined instruction spaces or labels by using LLM-generated relabeling, significantly increasing flexibility and open-ended task coverage.

\noindent \textbf{Leveraging Large Language Models in Reinforcement Learning.} Recent work has increasingly explored the use of pretrained Large Language Models (LLMs) to assist various aspects of RL tasks, including semantic reward shaping~\citep{xie2023text2reward,ma2023eureka}, task decomposition~\citep{huang2022language,lin2022inferring}, and high-level action guidance~\citep{du2023guiding,yao2023react,fan2022minedojo,wang2025m3hf}. Such methods exploit the reasoning power and semantic understanding capabilities of LLMs to enrich RL policy learning, although they often focus on forward generation of guidance or rewards from human-written task descriptions. By contrast, our proposed approach uniquely applies LLMs in a retrospective fashion: generating meaningful instruction labels from collected trajectory interactions, thus enabling effective utilization of unsuccessful episodes and improving sample efficiency and generalization across open-ended natural language instructions.

\section{Preliminaries}

\noindent \textbf{Markov Decision Process (MDP).} A Markov Decision Process (MDP) provides a mathematical framework for modeling sequential decision-making problems \citep{puterman1994,sutton2018reinforcement}. Formally, an MDP is defined as a tuple \((\mathcal{S}, \mathcal{A}, T, R, \gamma)\), where \(\mathcal{S}\) is the state space, \(\mathcal{A}\) is the action space, \(T: \mathcal{S} \times \mathcal{A} \to \mathcal{S}\) describes the state transition dynamics, \(R: \mathcal{S} \times \mathcal{A} \to \mathbb{R}\) is the reward function, and \(\gamma \in [0,1)\) is the discount factor. The goal of an agent interacting with the MDP is to find a policy \(\pi: \mathcal{S} \to \mathcal{A}\) that maximizes the expected cumulative discounted reward:
\begin{equation}
    J(\pi) = \mathbb{E}_{s_0 \sim p(s_0), a_t \sim \pi(s_t)} \left[ \sum_{t=0}^{\infty} \gamma^t R(s_t, a_t) \right].
\end{equation}

\noindent \textbf{Hindsight Experience Replay (HER).} Reinforcement learning algorithms often struggle in sparse reward environments, as successful experiences occur infrequently. Hindsight Experience Replay (HER) \citep{andrychowicz2017hindsight} addresses this challenge by enabling agents to learn effectively from failures through goal relabeling. Specifically, HER retrospectively reinterprets unsuccessful episodes by setting the goals to states the agent actually achieved later in the trajectory.

Formally, given an observed trajectory \(\tau = \{(s_t, a_t, r_t, s_{t+1}, g)\}_{t=0}^{T-1}\) associated with an original goal \(g\), HER selects an achieved future state \(s_{t'}\) (where \(t' > t\)) from the trajectory and relabels the original goal as \(g' = s_{t'}\). The reward for the relabeled trajectory is then recomputed accordingly:
\begin{equation}
    r_t' = R(s_t, a_t, g').
\end{equation}

By using achieved states as hindsight goals, HER significantly enhances the agent's ability to learn efficiently from sparse rewards, effectively converting unsuccessful episodes into valuable learning experiences.

\section{Method}

We propose a novel approach leveraging hindsight instruction relabeling guided by large language models (LLMs) to efficiently train generalizable instruction-conditioned reinforcement learning (RL) policies. Unlike traditional hindsight experience replay techniques~\citep{andrychowicz2017hindsight}, which reuse visited states to generate goals, our method synthesizes diverse free-form textual instructions directly from collected trajectories. This synthesis allows agents to learn from a richer set of instructions without requiring domain-specific knowledge, thereby facilitating effective training in environments characterized by sparse and semantic reward signals.

We formalize instruction-following RL tasks as an MDP extended with an explicit instruction space $(\mathcal{S}, \mathcal{A}, \mathcal{I}, R, T, \gamma)$, where $\mathcal{S}$ is the state space, $\mathcal{A}$ the action space, and $\mathcal{I}$ is the (potentially unbounded) textual instruction space expressed in natural language.

At each timestep $t$, the agent observes a state $s_t \in \mathcal{S}$ and an instruction $i \in \mathcal{I}$. It then selects an action $a_t \in \mathcal{A}$ according to a conditional policy $a_t \sim \pi_{\theta}(a_t \mid s_t, f_{\text{instr}}(i))$,
where $f_{\text{instr}}(\cdot)$ is an embedding function encoding textual instructions into embedding vectors. In our setup, we assume access to a pretrained instruction embedding encoder $f_{\text{instr}}$, such as SBERT~\citep{reimers2019sentence}.

Subsequently, the environment transitions to the next state $s_{t+1} \sim T(s_t, a_t)$ and produces a reward explicitly conditioned on the current instruction $r_t = R(s_t, a_t, i)$.

We assume that the ground truth reward function $R$ is binary (with $R(s_t, a_t, i)=1$ indicating successful completion and $0$ otherwise), but typically not acessible for arbitrary instructions during training. To address this challenge, we use hindsight instruction relabeling with automatically generated instructions from LLMs, effectively synthesizing surrogate binary reward signals based on collected trajectories to enable efficient learning.

Finally, we express our overall training and evaluation objective through a statistical mapping $f$, aggregating individual instruction-conditioned returns (or success indicators) into a single scalar metric:
\begin{equation}
\max_{\theta} f\left(\mathbb{E}_{i \sim \mathcal{I}}\left[G(i,\pi_{\theta})\right]\right),
\label{eq:objective_general}
\end{equation}
where $G(i,\pi_{\theta})$ denotes the instruction-conditioned cumulative reward or binary achievement indicator. Specific instantiations of the aggregation function $f$ include mean cumulative reward across instructions (expected return), success rate, and aggregated achievement score. We describe and evaluate these metrics concretely in Section~\ref{sec:experiments}.

\begin{algorithm}[t]
    \caption{\textbf{OIR}: LLM-Guided Hindsight Instruction Relabeling}
    \label{alg:main_alg}
    \small
    \begin{algorithmic}[1]
        \Require pretrained LLM $\mathcal{L}$, encoders $f_{\mathrm{state}},f_{\mathrm{instr}}$, off-policy RL algorithm $\texttt{alg}$
        \State Initialise instruction buffer $\mathcal{B}$ and $E$ parallel environments
        \For{each iteration}
            \State $\mathcal{D} \gets \emptyset$
            \State Collect trajectories $\{\tau_e\}_{e=1}^{E}$ with policy $\pi_{\theta}$
            \For{$e = 1 \;\textbf{to}\; E$}
                \State $\texttt{prompt}_e \gets h(\tau_e)$ \Comment{build LLM prompt}
                \For{$k = 1 \;\textbf{to}\; K$}
                    \State $i'_{e,k} \sim \mathcal{L}(\texttt{prompt}_e)$ \Comment{generate candidate}
                    \State $r^{\text{cand}}_{e,k} \gets R\!\left(i'_{e,k},\tau_e\right)$ \Comment{cosine-similarity reward, Eq.\;\ref{eq:rew}}
                    \State $\mathcal{D} \gets \mathcal{D} \cup \{(\tau_e,i'_{e,k},r^{\text{cand}}_{e,k})\}$  %
                \EndFor
            \EndFor
            \State \textbf{Update} $\mathcal{B}$ with all the relabeled instructions $\{i'_{e,k}\mid e\in[E], k\in[K]\}$ using Section~\ref{sec:instruction-buffer}
            \State Update $\left(\theta,\psi\right)$ using batch $\mathcal{D}$ with $\texttt{alg}$
            \State Reset finished environments with instructions sampled uniformly from $\mathcal{B}$
        \EndFor
    \end{algorithmic}
\end{algorithm}

Detailed in Algorithm~\ref{alg:main_alg}, our method is composed of: (1) the instruction generation and relabeling procedure; (2) reward and episode termination assignment based on the new instructions; and (3) the prioritized instruction buffer used to efficiently manage the instructions utilized during rollouts.

\subsection{Trajectory Collection and Instruction Relabeling via LLMs}
\label{sec:llm_relabel}

During training, we concurrently deploy $E$ parallel instances of the environment, each collecting trajectories under policy $\pi_{\theta}(\cdot\mid o_t, i_t)$. We denote the set of collected trajectories as $\{\tau_e\}_{e=1}^{E}$, where each trajectory is a sequence consisting of state-action pairs $\tau_e = \{(s^{e}_t, a^{e}_t)\}_{t=0}^{T_e}$.

To generate meaningful instructions from these trajectories, we leverage the capacity of large language models to perform semantic reasoning over textual descriptions of observed interactions. Concretely, we first convert each trajectory $\tau_e$ into an interpretable textual format suitable for prompting a pretrained LLM $\mathcal{L}$. Formally, given a trajectory $\tau_e$, we construct a textual prompt structured temporally as follows:

\vspace{2mm}
\noindent\fcolorbox{blue}{blue!10}{
\parbox{\linewidth}{
\textcolor{blue!70!black}{\textbf{Prompt:}}\\[2pt]
\texttt{What instruction is this trajectory following?}\\[4pt]
\texttt{timestep 0: {textual observation \{\{$o_0^e$\}\}}, agent takes action \{\{$a_0^e$\}\}}\\[2pt]
\texttt{timestep 1: {textual observation \{\{$o_1^e$\}\}}, agent takes action \{\{$a_1^e$\}\}}\\[2pt]
\texttt{...}\\[2pt]
\texttt{timestep $T_e$: {textual observation \{\{$o_{T_e}^e$\}\}}, agent takes action \{\{$a_{T_e}^e$\}\}}\\[2pt]
}}
\vspace{2mm}

Using such temporally structured prompts, the LLM returns plausible instructions corresponding to the actions of the trajectory. In general, given the language model $\mathcal{L}$, a trajectory-based textual observation transformation function $h$, and hyperparameter $K$ controlling instruction quantity and diversity, the instruction generation process is given by:
\begin{equation}
\{ i_{e,k} \}_{k=1}^{K} \sim \mathcal{L}\left(\texttt{prompt}(\tau_e)\right),
\end{equation}
with $\texttt{prompt}$ being the prompt template shown above.

Thus, each trajectory yields a set of $K$ candidate instructions suitable for the corresponding behavior. These synthetically generated instructions are subsequently used to generate synthetic reward and termination signals.

\noindent \textbf{Relabeling Failed Trajectories.}
An important property of our method is its ability to explicitly leverage and reinterpret trajectories that fail under their originally assigned instructions. Even when the agent does not successfully achieve the intended task during data collection, the hindsight relabeling procedure can still extract meaningful learning signals by deriving alternative instructions associated with the same behavioral sequence.

To formalize this concept clearly, we first define an oracle instruction-conditioned policy as:
\begin{equation}
\pi^{\textrm{oracle}} = \arg\max_{\pi} \mathbb{E}_{T,s_0}\left[ \sum_{t=0}^{\infty} \gamma^t R(s_t, a_t, i) \mid a_t \sim \pi(\cdot \mid s_t, f_{\textrm{instr}}(i)),\ s_0 \sim p(s_0),\ s_{t+1} \sim T(\cdot \mid s_t, a_t) \right],
\end{equation}
where the expectation is taken over the environment transition dynamics $T$ and the initial state distribution $p(s_0)$. Under this policy, the expected future cumulative reward of a state-instruction pair $(s, i)$ defines the value function:
\begin{equation}
V^{\pi^{\textrm{oracle}}}_i(s) = \mathbb{E}_{T,s_0}\left[ \sum_{t=0}^{\infty} \gamma^t R(s_t, a_t, i) \mid s_0 = s, a_t \sim \pi^{\textrm{oracle}}(\cdot \mid s_t, f_{\textrm{instr}}(i)),\ s_{t+1} \sim T(\cdot \mid s_t, a_t) \right].
\end{equation}

Given this formalism, we propose the following criterion for identifying effective alternative instructions from the LLM-generated candidates. Specifically, a candidate instruction $i'$ proposed at state-action pair $(s_t,a_t)$ for the observed trajectory is considered effective if assigning this instruction leads to a strictly greater value than the original sampled instruction $i_{\textrm{orig}}$:
\begin{equation}
V^{\pi^{\textrm{oracle}}}_{i'}(s_{t+1}) > V^{\pi^{\textrm{oracle}}}_{i_{\textrm{orig}}}(s_{t+1}).
\end{equation}

Practically, as directly accessing the oracle value function is infeasible, we rely on the language model to generate candidate alternative instructions based on the observed trajectory. These candidates are then subject to further filtering or selection to evaluate their effectiveness.

Notably, LLMs can sometimes generate inaccurate or misleading instruction candidates, either due to limited access to environment-specific information or challenges in handling long observation trajectories. To mitigate this, we incorporate rule-based instructions as an additional source of supervision. This also motivates the subsequent selection and validation steps, which are designed to ensure the quality of the relabeled instructions.

\subsection{Reward Definition and Episode Termination Criterion}
\label{sec:reward_termination}

To flexibly handle diverse, free-form instructions, we utilize an embedding-based reward function defined by semantic cosine similarity following ELLM~\citep{gallici2024simplifying}. Let $f_\text{state}(o_t)$ denote the embedding of transition $(o_t, a_t, o_{t+1})$ and $f_\text{instr}(i)$ the embedding of instruction $i$. At each time step $t$, the reward for instruction completion is given by
\begin{equation}\label{eq:rew}
    r_t(o_t, i) = \operatorname{cosim}(f_\text{state}(o_t, a_t, o_{t+1}), \, f_\text{instr}(i)),
\end{equation}
where the cosine similarity is defined as
$
    \operatorname{cosim}(\mathbf{a}, \mathbf{b}) = \frac{\mathbf{a} \cdot \mathbf{b}}{ \| \mathbf{a} \| \, \| \mathbf{b} \| }.$ An episode is deemed successful the first time the reward exceeds a predefined threshold $\delta$,  $ r_t(o_t, i) > \delta $, 
at which point the episode is marked as terminated.

\subsection{Prioritized Instruction Replay Buffer} \label{sec:instruction-buffer}
To handle the enlarged instruction set while keeping memory bounded, we adopt an 
eviction--rather than a sampling--based replay strategy inspired by Prioritized 
Level Replay (PLR)~\citep{jiang2021prioritized}.  
For every instruction $i$ we maintain its empirical mean return 
$\bar{R}(i)$, which are used only to \emph{order} new instructions before they are inserted; afterwards, instructions are drawn 
uniformly at random from the buffer.

\noindent \textbf{Priority ordering.} Each incoming instruction is first assigned to one of three mutually exclusive categories,
\begin{equation}
\mathrm{Status}(i) =
\begin{cases}
0 = \text{Learning--boundary}, & \tau_{\mathrm{low}} < \bar{R}(i) \le \tau_{\mathrm{high}},\\
1 = \text{Failing}, & \bar{R}(i) \le \tau_{\mathrm{low}},\\
2 = \text{Mastered}, & \bar{R}(i) > \tau_{\mathrm{high}}.
\end{cases}
\end{equation}
where $0 < \tau_{\text{low}} < \tau_{\text{high}}$ are fixed thresholds.
Instructions in the learning–boundary category are considered most valuable
because they reveal the agent’s current frontier of competence.  
Within each category we break ties with $\bar{R}(i)$, 
preferring instructions that have been seen fewer times.  
Formally, we sort all candidates lexicographically by the tuple  
\begin{equation}
\bigl(\mathrm{Status}(i),\; \bar{R}(i)\bigr), \qquad 
\mathrm{Status}(i) \in \{0,1,2\}.
\end{equation}
Smaller tuples indicate higher eviction priority.

\noindent \textbf{Round-robin eviction.} The replay buffer is a fixed-size circular array.
After sorting, we iterate through the ordered list of new instructions and
insert them sequentially, overwriting existing entries in 
round-robin fashion.  In this way the buffer always contains the most 
recent instructions from the top of the priority list while still retaining a
mixture of older tasks.

\noindent \textbf{Uniform sampling at episode termination.} Whenever an environment reaches the end of an episode, the next instruction
is selected uniformly at random from the buffer.  
Because the buffer composition is itself prioritized, this simple sampling
rule suffices to focus training on tasks close to the agent’s learning
boundary while still providing occasional exposure to harder and already
mastered instructions.

This eviction-based replay mechanism balances exploration (by continually
introducing under-sampled or failing tasks) and exploitation (by frequently
retaining learning-boundary tasks), leading to faster and more robust policy
improvement without the need for complicated probability computations or
importance-sampling corrections.

\section{Experiments}
\label{sec:experiments}

We empirically evaluate our proposed method to assess its capability in learning instruction-conditioned RL policies, specifically addressing three central research questions.
\begin{itemize}
    \item \textbf{RQ1 (Efficiency):} Does the integration of LLM-guided hindsight instruction relabeling improve training efficiency compared to baseline methods?
    \item \textbf{RQ2 (Generalization):} Can our method generalize beyond training instructions and successfully handle previously unseen instructions through semantic supervision obtained from LLM-generated instruction relabeling?
    \item \textbf{RQ3 (Diversity):} Does our approach increase the semantic diversity and coverage of instructions compared to baseline methods?
\end{itemize}

\begin{figure}[t]
    \centering
    \includegraphics[width=0.93\linewidth]{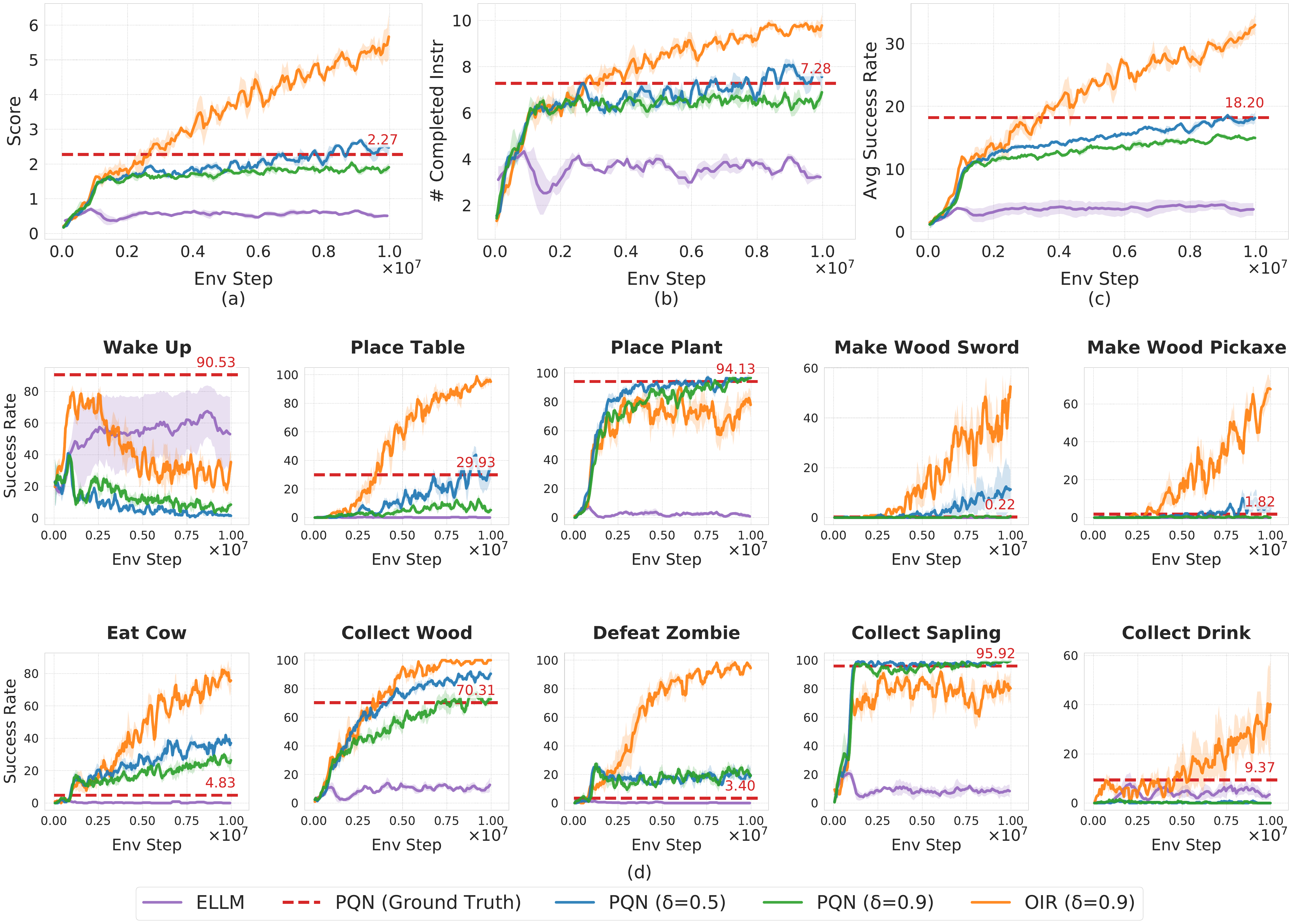}
    \caption{(a-c) Performance comparison of OIR against baseline methods measured by (a) aggregate score, (b) number of completed instructions, and (c) average success rate across all original instructions. (d) Success rates of OIR compared to baselines for individual instructions. OIR consistently outperforms baseline methods across all evaluation metrics and nearly all individual tasks. Results are averaged over three random seeds, with shaded areas representing standard errors.}
    \label{fig:main-exp}
    \vspace{-0.8cm}
\end{figure}

\noindent \textbf{Environment.} We adopt the Craftax-Classic environment~\citep{matthews2024craftax}, an open-ended reinforcement learning benchmark that provides diverse, procedurally-generated environments explicitly characterized by textual instructions (\textit{achievements}), which makes it particularly suitable for evaluating instruction-conditioned RL methods. Craftax-Classic originally offers sparse achievement notifications linked explicitly to environment-defined tasks (e.g., \textit{``collect wood''}). However, we completely remove all built-in rewards and environment-provided achievement signals during training, making it necessary for the agent to rely exclusively on its own to explore the open-ended environment.

\noindent \textbf{Training Protocol.}
We conduct all experiments using \textit{parallelised $Q$-network} (PQN)~\citep{gallici2024simplifying} as our base RL algorithm. PQN enables efficient sampling across multiple parallel environment instances, allowing batched interaction and large-scale querying of the LLM for instruction generation. Throughout training, the agent does not receive reward signals from the environment. For fair and direct comparison, all evaluated baselines are implemented using the same PQN algorithm backbone.

\noindent \textbf{Evaluation Protocol.}
For the evaluation (conducted periodically during training), agent policies are assessed on a pre-defined suite of instructions covering three types:
\begin{enumerate}
    \item \textbf{Original Instructions}: Standard Craftax-achievements defined by the environment (e.g., \texttt{collect wood}).
    \item \textbf{Simple Variant Instructions}: Three linguistic variations per original instruction, assessing agent robustness to superficial textual differences (e.g., \texttt{collect wood} $\rightarrow$ \texttt{pick up logs from the ground.}).
    \item \textbf{Complex Variant Instructions}: Three semantically enriched and compositional variants testing deeper instruction-following capabilities (e.g., \texttt{Your inventory requires wood; chop down several trees.}).
\end{enumerate}
The evaluation provides external ground-truth success signals given by the environment. These signals are used \textit{only at evaluation} to measure true instruction-following capabilities.

\begin{figure}[t]
    \centering
    \includegraphics[width=0.87\linewidth]{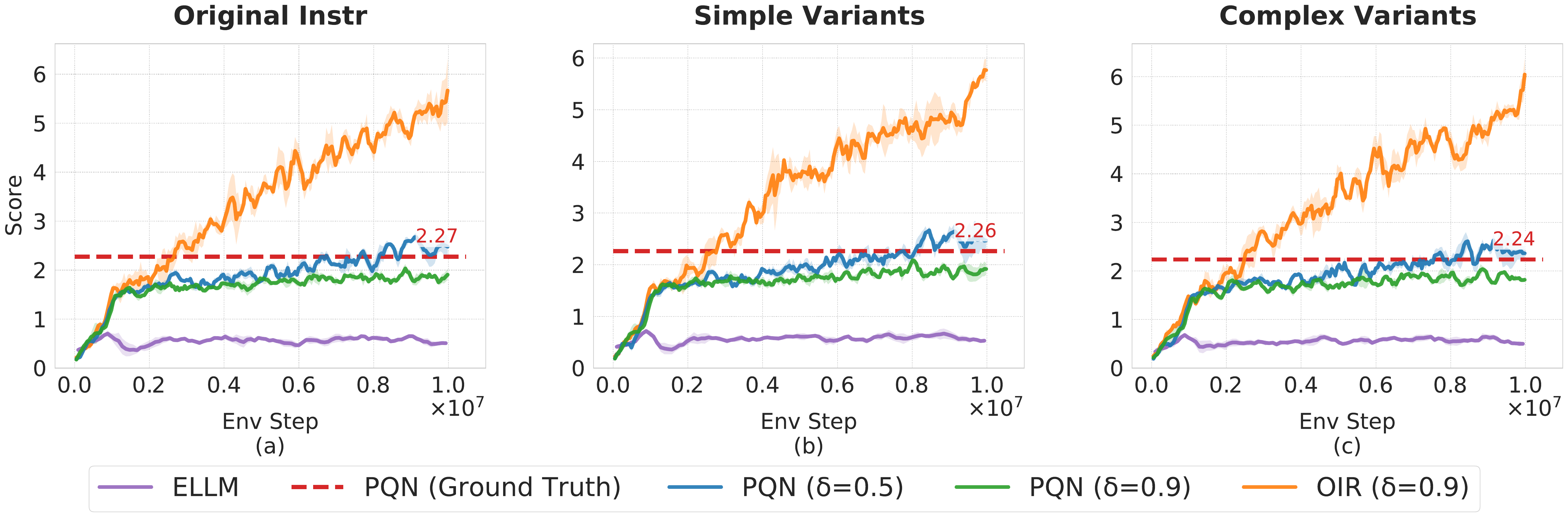}
    \caption{Generalization performance of OIR compared to baseline methods measured by aggregated score evaluation on (a) original instructions, (b) simple variant instructions, and (c) complex variant instructions. OIR demonstrates superior generalization capabilities, significantly outperforming baseline methods across all three variants. Results are averaged across three random seeds, with shaded areas denoting standard errors.}
    \label{fig:generalization}
    \vspace{-0.5cm}
\end{figure}

\noindent \textbf{Evaluation Metrics.}
To quantitatively evaluate the learned instruction-following behavior, we employ three primary metrics:
\begin{itemize}
    \item \textbf{Mean Success Rate}: Arithmetic average of the success rates across all individual instructions.
    \item \textbf{Mean Number of Completed Instructions}: The count of distinct instructions for which the policy achieves a success rate greater than zero.
    \item \textbf{Aggregate Score}: Calculated as $\exp\big(\frac{1}{N} \sum_{i=1}^{N} \log(1 + s_i)\big) - 1$, where $s_i$ denotes the success rate for instruction $i$. This scoring approach emphasizes improvements on instructions with lower success rates, rather than on those for which performance is already proficient.%
\end{itemize}

\noindent \textbf{Baselines.} We compare our proposed method against three representative baseline approaches. First, we adopt \textbf{PQN w/ cosine similarity reward}, which utilizes the PQN trained directly on the predefined \textit{Craftax} instructions, leveraging the same cosine similarity reward function as ours. Second, we consider \textbf{PQN w/ Ground-Truth Reward}, where PQN is trained on the ground-truth sparse binary achievement rewards provided by the environment (+1 upon the achievement of the instructed task, 0 otherwise); this baseline serves as an approximate performance upper bound, as it leverages the fully accurate reward. Finally, we include \textbf{ELLM (Exploring with LLMs)} \citep{du2023guiding}, where exploratory goals are generated via a LLM. For fairness, we adapt ELLM to the same PQN backbone as our method and remove any domain-specific heuristics and engineering.

\subsection{RQ1 (Efficiency)}

From Figures~\ref{fig:main-exp}(a)-(c), we observe that our method OIR surpasses all baselines across every aggregated evaluation metric.
Similarly, OIR demonstrates superior performance in terms of the number of completed instructions, achieving approximately $10$ completed tasks compared to fewer than four tasks for other baselines. This indicates that our method not only achieves better task-specific performance but also scales better across the diverse set of instructions. The overall average success rate across instructions (33.10\%) further confirms that OIR leads to consistently improved policy performance and learning efficiency compared to traditional PQN methods and ELLM. 

Figure~\ref{fig:main-exp}(d) further elucidates performance differences at the individual-instruction level. OIR learns challenging instructions, such as \texttt{"Defeat Zombie"} (87.78\%), \texttt{"Place Table"} (95.83\%), and \texttt{"Make Wood Pickaxe"} (76.59\%), which other baselines completely fail to master. Such hard instructions require multi-step exploration and purposeful behavior, clearly demonstrating that the semantic guidance from open-ended instructions generated by the LLM substantially improves the exploration and learning outcomes in sparse-reward scenarios.

However, it is important to recognize that OIR does not universally outperform other methods on every individual instruction. For instance, the instruction \texttt{"Wake Up"} consistently yields better performance with the ground-truth PQN baseline (90.53\%) compared to OIR. This reflects a trade-off inherent to our method's design: due to the fixed buffer capacity of relabeled instructions, the agent may gradually reduce proficiency on instructions that occur infrequently or lack semantic relevance to a diverse set of other instructions. Such instructions are less likely to be frequently sampled during policy updates and can decrease in performance over training. This trade-off, nonetheless, is balanced by the overall substantial improvement across a broader and richer set of instructions.

\begin{figure}[t]
    \centering
    \includegraphics[width=0.9\linewidth]{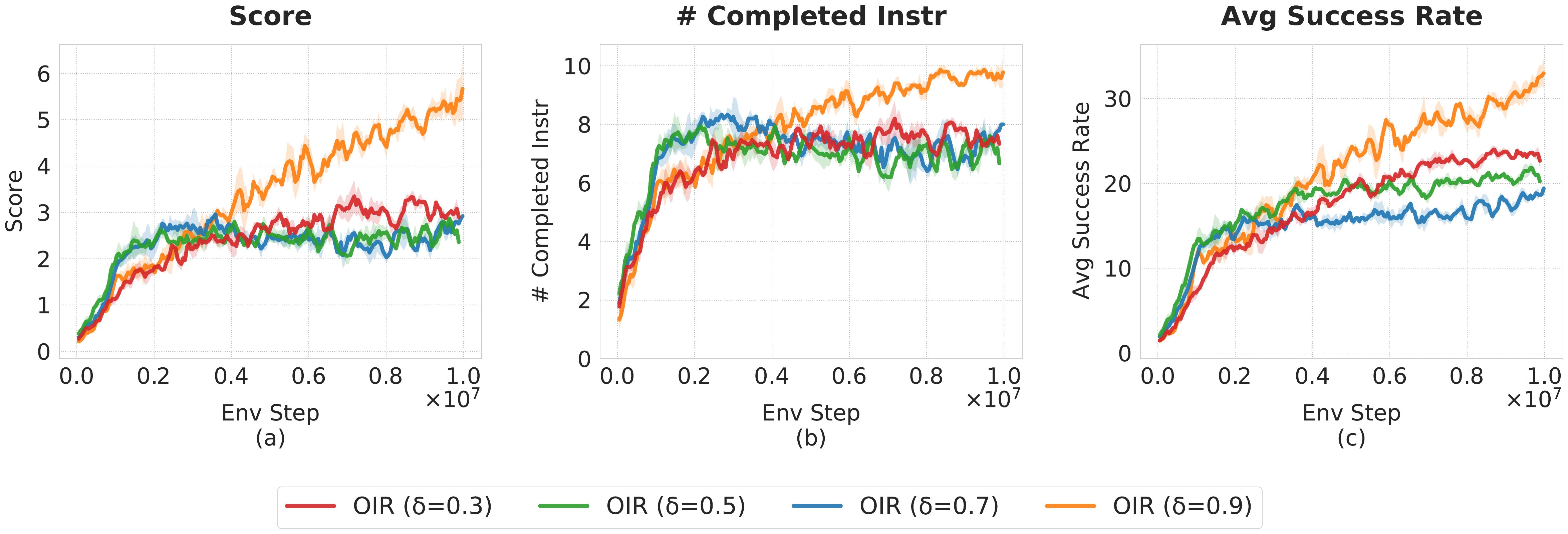}    \caption{Sensitivity of OIR to the different cosine-similarity threshold $\delta$}
    \label{fig:ablation}
    \vspace{-0.5cm}
\end{figure}

\subsection{RQ2 (Generalization)}

According to Figure~\ref{fig:generalization}, our OIR method consistently outperforms baselines across all instruction categories, including previously unseen variations. In contrast, baseline methods show nearly identical performance across original and previously unseen instruction variants, but at notably lower aggregate scores. Combined with the observation in Figure~\ref{fig:main-exp}(d) that these baselines learn to successfully complete only a small and fixed subset of instructions, we conclude that their apparently stable performance reflects an inability to learn nuanced instruction-following behaviors: instead, they execute shallow and repetitive actions that are independent of semantic instruction context. Overall, these results highlight the clear advantage and scalability of our OIR method in effectively generalizing instruction-following policies to handle diverse and previously unseen natural-language instruction variants.

\begin{wrapfigure}[14]{r}{0.36\textwidth}
\vspace{-1.5cm}
\begin{center}\includegraphics[width=0.95\linewidth]{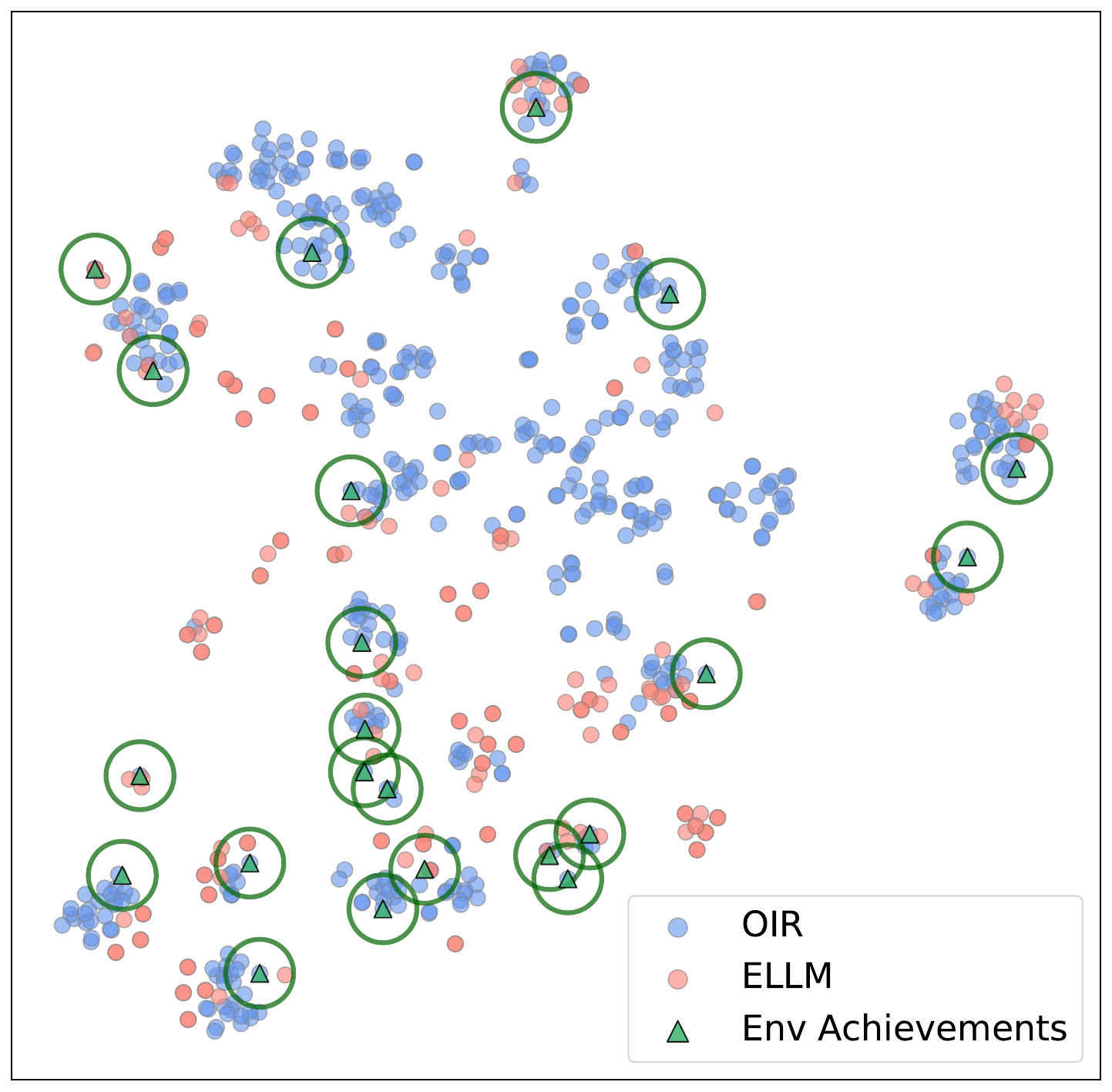}
        \caption{t-SNE visualization of semantic diversity of instructions generated by OIR compared to ELLM and environment-provided achievements.}
        \label{fig:vis}
    \end{center}
\end{wrapfigure}

\subsection{RQ3 (Diversity)}

To address RQ3, we visually interpret generated instructions in semantic embedding space using a two-dimensional t-SNE embedding in Figure~\ref{fig:vis}. Each data point corresponds to an individual instruction generated during policy training. We observe that OIR-generated instructions broadly populate a significantly larger and richer subregion of the embedding space, extending notably beyond the original Craftax achievements. In contrast, instructions from ELLM predominantly cluster near predefined achievements, suggesting restricted semantic variation and limited exploratory signals.

Critically, the greater semantic coverage of OIR is directly attributable to our hindsight instruction relabeling strategy. By retrospectively generating instructions from collected trajectories (including unsuccessful rollouts), OIR naturally provides a more diverse instruction distribution that semantically explains agent behaviors.

\noindent \textbf{Ablation.} Figure \ref{fig:ablation} indicates that our method's effectiveness varies considerably with the cosine-similarity threshold ($\delta$). Choosing a lower threshold value leads to more frequent and denser rewards initially, enabling faster early-stage learning. However, these dense signals may overly reward partial or incomplete task performance, thereby providing potentially misleading guidance and ultimately hindering higher-level learning. In contrast, a higher threshold, though initially resulting in sparser rewards and slower early convergence, ensures semantically clearer instructional signals.

\section{Conclusion}
\label{sec:conclusion}
We proposed OIR to learn instruction-following policies by leveraging open-ended instruction relabeling with large language models. Our method automatically generates instructions from collected trajectories, effectively reducing reliance on human annotation. Experiments on the Craftax environment demonstrated improved sample efficiency, the capability to master challenging instructions, and better semantic coverage over the instruction space compared to baseline methods.

However, our method depends on the quality of instructions generated by pretrained LLMs, potentially inheriting their biases or inaccuracies. Additionally, the limited capacity of our instruction buffer can lead to the forgetting of infrequently sampled tasks. Future work should explore improved buffer management, instruction filtering, and human-in-the-loop verification to enhance robustness and applicability.

\section{Acknowledgements}
This work was supported in part by NSF grant 1S-2046640 (CAREER) and the Engineering and Physical Sciences Research Council [grant number EP/Y003187/1; UKRI849].

\bibliography{ref}
\bibliographystyle{unsrtnat}

\newpage

\newpage
\appendix

\section{Environment Details}
We adopt the codebase for Craftax~\citep{matthews2024craftax} available at \url{https://github.com/MichaelTMatthews/Craftax}.

\subsection{State Space}
\begin{itemize}
    \item \textbf{World grid}: discrete $H{\times}W$ array of block IDs.
    \item \textbf{Mobs}: position, health, cooldown and alive-mask for up to a fixed budget of \emph{zombies}, \emph{cows}, \emph{skeletons}, \emph{arrows}, and \emph{growing plants}.
    \item \textbf{Player avatar}: position $(x,y)$, facing direction, health, food, drink, energy, fatigue, thirst, sleep flag.
    \item \textbf{Inventory}: integer counts (0-9) for \{\textit{wood, stone, coal, iron, diamond, sapling, wood/stone/iron pickaxe, wood/stone/iron sword}\}.
    \item \textbf{Achievements}: per-task Boolean flags (defeat, craft, collect, etc.).
    \item \textbf{Time/light}: global timestep and continuous light level in $[0,1]$.
\end{itemize}

\subsection{Observation Space}
Agent receives a fixed-length vector obtained from a Chebyshev radius $R{=}3$ ($7{\times}7$ window) centered on the player.  Each tile is one-hot-encoded and concatenated with:
\begin{itemize}
    \item counts of each mob type at distances $1,\dots,R$,
    \item the ID of the block (or mob) directly in front of the player,
    \item the full inventory vector,
    \item the four vital statistics (health, food, drink, energy).
\end{itemize}
All features are normalised to $\,[0,1]$ and packed into a \texttt{Box} in $\mathbb{R}^d$.

\subsection{Textual Observation}
Each step also supplies a single natural-language sentence constructed as  
\texttt{``Facing: \ldots; Nearby: \ldots; Inventory: \ldots; Status: \ldots''}  
where
\begin{enumerate}
    \item \textbf{Facing} names the object or mob on the tile ahead,
    \item \textbf{Nearby} lists non-trivial blocks/mobs in the $7{\times}7$ window, grouped by distance and alphabetised within each group,
    \item \textbf{Inventory} enumerates items with non-zero count,
    \item \textbf{Status} reports health, food, drink, energy and whether the avatar is sleeping.
\end{enumerate}
The string is tokenised, padded to a fixed length, and embedded once per step.

\subsection{Action Space}
Discrete set of $|\mathcal{A}|{=}17$ actions:
\begin{center}
\begin{tabular}{@{}ll@{}}
0 & \textsc{noop} \\ \midrule
1-4 & \textsc{left}, \textsc{right}, \textsc{up}, \textsc{down} \\ \midrule
5 & \textsc{do / interact} \\ 
6 & \textsc{sleep} \\ \midrule
7-10 & \textsc{place\_stone}, \textsc{place\_table}, \textsc{place\_furnace}, \textsc{place\_plant} \\ \midrule
11-13 & \textsc{make\_wood\_pickaxe}, \textsc{make\_stone\_pickaxe}, \textsc{make\_iron\_pickaxe} \\ 
14-16 & \textsc{make\_wood\_sword}, \textsc{make\_stone\_sword}, \textsc{make\_iron\_sword}
\end{tabular}
\end{center}
These cover movement, interaction, resting, block placement, and crafting. To make the text Description more informative, we replace the generic \textsc{do / interact} with the actual object in front of the agent, e.g., grass.

\subsection{Testing Protocol Details}

\paragraph{Original Instructions}

We adopt the following 22 original environment-provided instructions as the ground truth:

\begin{itemize}[noitemsep,nolistsep,label=--]
    \item collect wood
    \item place table
    \item eat cow
    \item collect sapling
    \item collect drink
    \item make wooden pickaxe
    \item make wooden sword
    \item place plant
    \item defeat zombie
    \item collect stone
    \item place stone
    \item eat plant
    \item defeat skeleton
    \item make stone pickaxe
    \item make stone sword
    \item wake up
    \item place furnace
    \item collect coal
    \item collect iron
    \item collect diamond
    \item make iron pickaxe
    \item make iron sword
\end{itemize}

\paragraph{Simple \& Complex Variants}

Below are examples illustrating simple and complex variants for $5$ instructions:

\begin{itemize}[leftmargin=*, itemsep=1em]

    \item \textbf{Place Table}
    \begin{itemize}[label=--, noitemsep, leftmargin=*]
      \item \textit{Simple:}
        \begin{itemize}[label=$\circ$, noitemsep]
          \item A construction bench is needed; deploy one.
          \item The floor is a good spot for a new making-area.
          \item Time to set up the workbench.
        \end{itemize}
      \item \textit{Complex:}
        \begin{itemize}[label=$\circ$, noitemsep]
          \item Find a solid spot and deploy the construction apparatus there.
          \item Ready the zone for making things by placing the special surface.
          \item The workspace needs a crafting implement; position it.
        \end{itemize}
    \end{itemize}

    \item \textbf{Eat Cow}
    \begin{itemize}[label=--, noitemsep, leftmargin=*]
      \item \textit{Simple:}
        \begin{itemize}[label=$\circ$, noitemsep]
          \item Restore your energy by consuming cooked animal flesh.
          \item Meat from the beast is on the menu.
          \item Prepared creature meat should be eaten.
        \end{itemize}
      \item \textit{Complex:}
        \begin{itemize}[label=$\circ$, noitemsep]
          \item To stave off hunger, prepare and then consume the animal parts.
          \item Sustenance can be gained from the cooked beast meat.
          \item Replenish your energy by ingesting the prepared animal tissue.
        \end{itemize}
    \end{itemize}

    \item \textbf{Collect Sapling}
    \begin{itemize}[label=--, noitemsep, leftmargin=*]
      \item \textit{Simple:}
        \begin{itemize}[label=$\circ$, noitemsep]
          \item Young trees are available; gather one.
          \item A tree sprout is on the ground, pick it up.
          \item Procure a small tree for future planting.
        \end{itemize}
      \item \textit{Complex:}
        \begin{itemize}[label=$\circ$, noitemsep]
          \item For replanting efforts, obtain some juvenile trees.
          \item Tree seedlings must be collected and then stored.
          \item Environmental small tree starts are yours for the taking.
        \end{itemize}
    \end{itemize}

    \item \textbf{Collect Drink}
    \begin{itemize}[label=--, noitemsep, leftmargin=*]
      \item \textit{Simple:}
        \begin{itemize}[label=$\circ$, noitemsep]
          \item Clear liquid is present; get some.
          \item A beverage is available for pickup.
          \item H\textsubscript{2}O for later consumption should be gathered.
        \end{itemize}
      \item \textit{Complex:}
        \begin{itemize}[label=$\circ$, noitemsep]
          \item Ensure you have drinking liquid by filling a container.
          \item Potable water needs to be collected and bottled.
          \item From a nearby source, gather fluid for later use.
        \end{itemize}
    \end{itemize}

    \item \textbf{Make Wood Pickaxe}
    \begin{itemize}[label=--, noitemsep, leftmargin=*]
      \item \textit{Simple:}
        \begin{itemize}[label=$\circ$, noitemsep]
          \item A digging tool from tree material is the goal.
          \item Create a mining implement using wood.
          \item Lumber can be fashioned into a digging utensil.
        \end{itemize}
      \item \textit{Complex:}
        \begin{itemize}[label=$\circ$, noitemsep]
          \item The crafting table is where a digging implement of wood is made.
          \item Combine tree-based planks and sticks for a new mining tool.
          \item Construct a digging utensil; lumber components are required.
        \end{itemize}
    \end{itemize}

\end{itemize}

\section{Implementation Details}

\subsection{Algorithm Pseudocode}
We present the detailed version of the pseudocode as in Algorithm~\ref{alg:main_alg}.

\begin{algorithm}[t]
    \caption{\textbf{OIR}: LLM-Guided Hindsight Instruction Relabeling (detailed version)}
    \label{alg:long_alg}
    \small
    \definecolor{inputbg}{RGB}{235,245,255}
    \definecolor{collectbg}{RGB}{255,245,235}
    \definecolor{relbg}{RGB}{255,235,245}
    \definecolor{updatebg}{RGB}{235,255,255}
    \definecolor{currbg}{RGB}{245,235,255}
    \begin{algorithmic}[1]
    \Statex\colorbox{inputbg}{\parbox{\dimexpr\linewidth-2\fboxsep}{\textbf{Inputs}}}
        \Require pretrained LLM $\mathcal{L}$; state encoder $f_{\mathrm{state}}$; instruction encoder $f_{\mathrm{instr}}$
        \Require off-policy RL algorithm \texttt{alg} (parameters $\theta,\psi$)
        \Require \#~LLM candidates $K$; instruction buffer capacity $B_{\max}$; \#~parallel envs $E$
        \State Initialise instruction buffer $\mathcal{B}\leftarrow\emptyset$; launch $\{\texttt{Env}_{e}\}_{e=1}^{E}$
    \For{iteration $=1,2,\dots$}
    \Statex\colorbox{collectbg}{\parbox{\dimexpr\linewidth-2\fboxsep}{\textbf{(1) Trajectory Collection \& LLM Generation}}}
        \State $\mathcal{D}\gets\emptyset$
        \ForAll{envs $e=1,\dots,E$ \textbf{in parallel}}
            \State Roll out trajectory $\tau_e=\{(s^e_t,a^e_t)\}_{t=0}^{T_e}$ with current policy $\pi_{\theta}$
            \State $\texttt{prompt}_e \gets h(\tau_e)$ \Comment{build LLM prompt}
            \For{$k=1$ \textbf{to} $K$}
                \State $i'_{e,k}\sim\mathcal{L}(\texttt{prompt}_e)$ \Comment{generate candidate instruction}
                \State $r^{\text{cand}}_{e,k}\gets R(i'_{e,k},\tau_e)$ \Comment{cosine-similarity reward, Eq.\;\ref{eq:rew}}
                \State $\mathcal{D}\gets\mathcal{D}\cup\{(\tau_e,i'_{e,k},r^{\text{cand}}_{e,k})\}$
            \EndFor
        \EndFor
    \Statex\colorbox{relbg}{\parbox{\dimexpr\linewidth-2\fboxsep}{\textbf{(2) Instruction-Buffer Maintenance}}}
        \State \textbf{Update} $\mathcal{B}$ with all relabeled instructions $\{\,i'_{e,k}\mid e\in[E],k\in[K]\}$ using Section~\ref{sec:instruction-buffer}
        \If{$|\mathcal{B}|>B_{\max}$}
            \State \textproc{EvictOldest}$(\mathcal{B},|\mathcal{B}|-B_{\max})$
        \EndIf
    \Statex\colorbox{updatebg}{\parbox{\dimexpr\linewidth-2\fboxsep}{\textbf{(3) Policy Update}}}
        \State Update $(\theta,\psi)$ with one step of \texttt{alg} on batch $\mathcal{D}$
    \Statex\colorbox{currbg}{\parbox{\dimexpr\linewidth-2\fboxsep}{\textbf{(4) Environment Reset / Curriculum}}}
        \ForAll{envs $e$ that have terminated}
            \State Sample $i\sim\textproc{Uniform}(\mathcal{B})$
            \State Reset $\texttt{Env}_{e}$ with instruction $i$
        \EndFor
    \EndFor
    \end{algorithmic}
\end{algorithm}

\subsection{Hyperparameters}

\begin{table}[h!]
    \caption{Hyperparameters used in \textit{Craftax-Classic}.}
    \label{tab:hyperparams}
    \centering
    \begin{tabular}{lccc}
        \toprule
        \textbf{Hyperparameter} & \textbf{PQN (all versions)} & \textbf{ELLM} & \textbf{OIR (ours)} \\
        \midrule
        Total Timesteps & $1\times10^{7}$ & $1\times10^{7}$ & $1\times10^{7}$ \\
        Total Timesteps (decay) & $1\times10^{7}$ & $1\times10^{7}$ & $1\times10^{7}$ \\
        Number of Environments ($N_{\text{env}}$) & 64 & 1024 & 64 \\
        Steps per Environment ($N_{\text{steps}}$) & 128 & 8 & 128 \\
        $\epsilon_{\text{start}}$ & 1.0 & 1.0 & 1.0 \\
        $\epsilon_{\text{finish}}$ & 0.1 & 0.1 & 0.1 \\
        $\epsilon$ Decay Ratio & 0.1 & 0.1 & 0.1 \\
        Number of Minibatches & 4 & 4 & 4 \\
        Number of Epochs & 8 & 8 & 8 \\
        Input Normalization & True & True & True \\
        Normalization Type & Layer Norm & Layer Norm & Layer Norm \\
        Hidden Size & 512 & 512 & 512 \\
        Number of Layers & 1 & 1 & 1 \\
        Number of RNN Layers & 1 & 1 & 1 \\
        Add Last Action & False & False & False \\
        Learning Rate & $1\times10^{-5}$ & $1\times10^{-5}$ & $1\times10^{-5}$ \\
        Max Gradient Norm & 0.5 & 0.5 & 0.5 \\
        Linear LR Decay & True & True & True \\
        Discount Factor ($\gamma$) & 0.99 & 0.99 & 0.99 \\
        GAE $\lambda$ & 0.5 & 0.5 & 0.5 \\
        Instruction Buffer Size & - & - & 10 \\
        Shaping Threshold & 0.5\&0.9 & 0.9 & 0.9 \\
        LLM & - & Qwen3-8B & Qwen3-8B \\
        \bottomrule
    \end{tabular}%
\end{table}

Here we present in Table~\ref{tab:hyperparams} the hyperparameter configurations used in all of our experiments for the baselines (PQN and ELLM) and our algorithm OIR.

\subsection{Training Resources \& Training Time}

Our experiments were conducted on a server running Ubuntu 24.04. The hardware configuration included Dual AMD EPYC 7453 28-core processors, providing a total of 112 threads. For accelerated computing, we utilized two NVIDIA A6000 GPUs and two NVIDIA A6000 Ada GPUs. The total computational resources comprised 112 CPU threads and 512GB of system memory. Each training run for OIR and ELLM required approximately 6-8 hours to complete.

\subsection{Baseline Implementation Details}
We adopt the official PQN implementation available at \url{https://github.com/mttga/purejaxql}. For ELLM, we closely follow the methodology presented in the original paper~\citep{du2023guiding}, carefully implementing ELLM on top of PQN as described in the official repository (\url{https://github.com/yuqingd/ellm}).

To ensure a fair comparison, we remove environment-specific customizations originally included in their implementation, such as task decomposition and action-space-related checks for goal completion. Additionally, we increase the number of parallel environments to $1024$, enabling batched calls to the LLM at each step. Without batching, the turnaround time would be prohibitively high.

\clearpage
\subsection{Prompts}
\label{sec:prompts}

\begin{gbox}[prompt:code_extract1]{OIR Prompt}
\textbf{Description in Crafter:}

You are an expert Hindsight Instruction Relabeler for Minecraft agents.

You are a strict instruction generator. Under no circumstances may your output contain any movement or exploration instructions. Do \textbf{NOT} use the words \texttt{"move"}, \texttt{"explore"}, \texttt{"navigate"}, \texttt{"go to"}, or any synonyms indicating changing position.

\textbf{Environment:}
\begin{itemize}
  \item \textbf{Resources:} wood (trees), stone, coal, iron, water, sapling (from grass)
  \item \textbf{Tools:} crafting table, furnace, pickaxe (wood, stone, iron), sword (wood, stone, iron)
  \item \textbf{Mobs:} cow, zombie, skeleton
\end{itemize}

\textbf{Principles} (max 2 each; mention an entity in every instruction):

\textbf{Mid-Level} (atomic, 1–2 steps)
\begin{itemize}
  \item e.g., \texttt{"collect wood from the tree"}, \texttt{"collect stone"}, \texttt{"make a stone pickaxe"},
  \texttt{"place crafting table"}, \texttt{"place furnace"}, \texttt{"attack zombie"}, \texttt{"attack skeleton"},
  \texttt{"drink the water"}, \texttt{"wake up"}
\end{itemize}

\textbf{High-Level} (multi-step, purposeful)
\begin{itemize}
  \item e.g., \texttt{"Make tools for collecting the iron"},
  \texttt{"Collect wood then place a table, finally make a wood pickaxe"}
\end{itemize}

\textbf{Task:}

Given only a trajectory segment, output:
\texttt{
{
  "Analysis": "...",
  "Completed Instructions": {
    "Mid-Level": ["...", "...", "...", "..."],
    "High-Level": ["...", "...", "...", "..."]
  }
}}

\textbf{Example:}

Segment:

\texttt{0–4: move to tree}

\texttt{5: chop tree}

\texttt{7–8: place crafting table}

\texttt{9–10: make wood pickaxe}

\texttt{11–20: sleep}

\texttt{20–22: wake up}

Your answer:
\texttt{
{
  "Analysis": "Chopped tree first, then set up crafting table, finally make wood pickaxe.",
  "Completed Instructions": {
    "Mid-Level": [
      "collect wood from tree",
      "place crafting table",
      "make a wood pickaxe",
      "sleep and wake up"
    ],
    "High-Level": [
      "Prepare to collect stone",
      "collect tools to mine stone and coal",
      "prepare all tools to collect stone, then make stone pickaxe"
    ]
  }
}}
\newline\newline
\textbf{\{\{trajectory\}\}}

\end{gbox}
\clearpage

\end{document}